\PassOptionsToPackage{svgnames,dvipsnames}{xcolor}
\documentclass[journal]{vgtc}                     

\onlineid{1272}

\vgtccategory{Research}

\vgtcpapertype{Analytics \& Decisions}

\title{Illuminating LLM Coding Agents: Visual Analytics for Deeper Understanding and Enhancement}

\author{%
  Junpeng Wang, Yuzhong Chen, Menghai Pan, Chin-Chia Michael Yeh, and Mahashweta Das
}

\authorfooter{
  \item
  	J. Wang, Y. Chen, M. Pan, C.-C. M. Yeh, M. Das are with Visa Research.
  	E-mail: \{junpenwa, yuzchen, menpan, miyeh, mahdas\}@visa.com

}

\abstract{%
Coding agents powered by large language models (LLMs) have gained traction for automating code generation through iterative problem-solving with minimal human involvement. Despite the emergence of various frameworks, e.g., LangChain, AutoML, and AIDE, ML scientists still struggle to effectively review and adjust the agents' coding process. The current approach of manually inspecting individual outputs is inefficient, making it difficult to track code evolution, compare coding iterations, and identify improvement opportunities. To address this challenge, we introduce a visual analytics system designed to enhance the examination of coding agent behaviors. Focusing on the AIDE framework, our system supports comparative analysis across three levels: (1) Code-Level Analysis, which reveals how the agent debugs and refines its code over iterations; (2) Process-Level Analysis, which contrasts different solution-seeking processes explored by the agent; and (3) LLM-Level Analysis, which highlights variations in coding behavior across different LLMs. By integrating these perspectives, our system enables ML scientists to gain a structured understanding of agent behaviors, facilitating more effective debugging and prompt engineering. Through case studies using coding agents to tackle popular Kaggle competitions, we demonstrate how our system provides valuable insights into the iterative coding process.
}


\teaser{
  \centering
  \includegraphics[width=\linewidth, alt={A view of a city with buildings peeking out of the clouds.}]{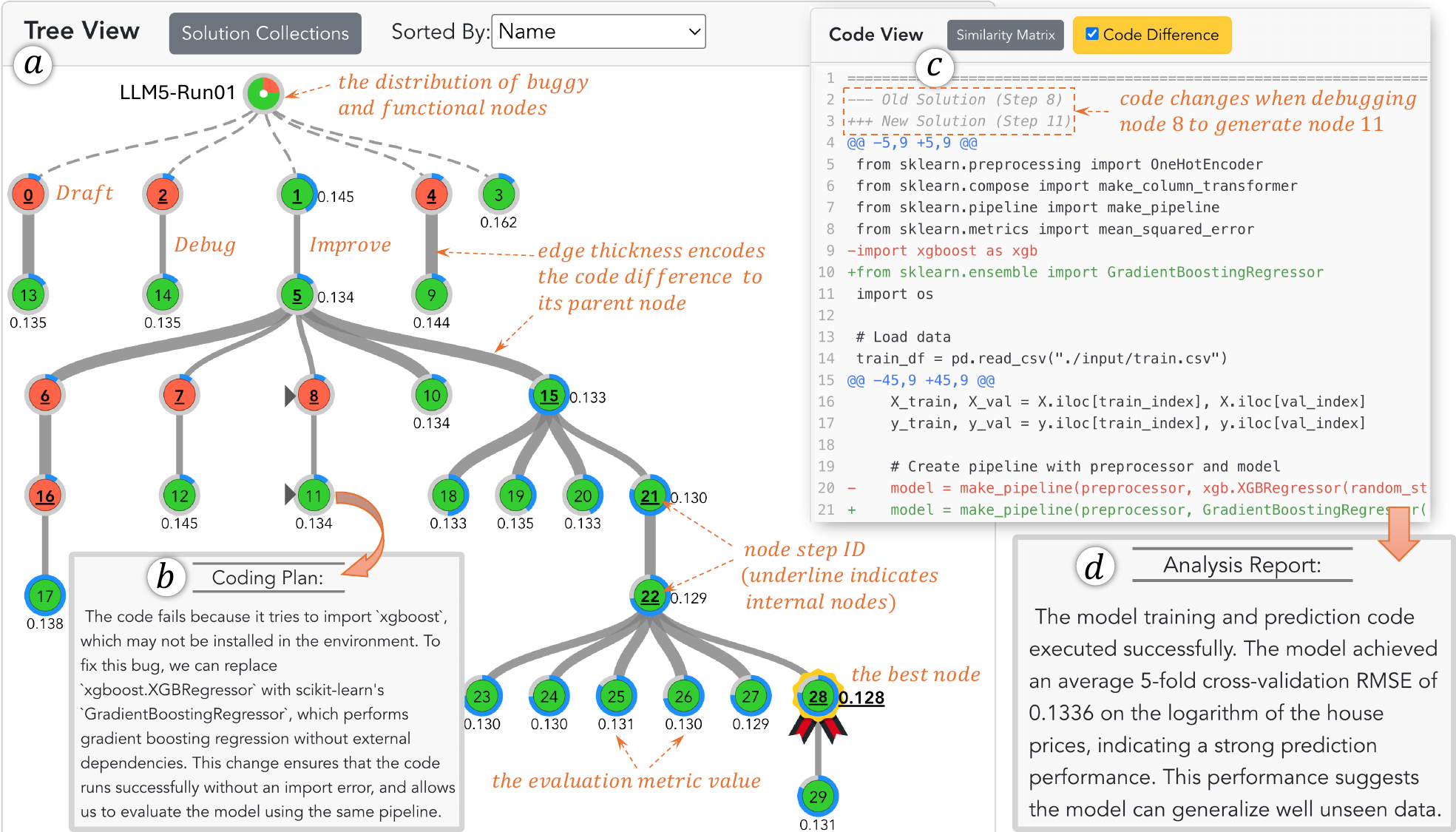}
  \vspace{-0.2in}
  \caption{Our visual analytics system uses an interactive tree (a) to demonstrate the iterative solution-seeking (coding) process of AIDE. The coding plan (b), generated code (c), and code analysis report (d), can be presented on demand while interacting with the tree.
  }
  \label{fig:system1}
}

\graphicspath{{figs/}{figures/}{pictures/}{images/}{./}} 

\usepackage{xcolor}
\usepackage{tabu}                      
\usepackage{booktabs}                  
\usepackage{lipsum}                    
\usepackage{mwe}                       
\usepackage{enumitem}
\usepackage{mathptmx}                  
\usepackage[most]{tcolorbox}

\newcommand{\treeview}{\textit{Tree View}}
\newcommand{\codeview}{\textit{Code View}}
\newcommand{\projectionview}{\textit{Projection View}}
\newcommand{\packageview}{\textit{Package View}}
\newcommand{\llmone}{\texttt{LLM1}} 
\newcommand{\llmtwo}{\texttt{LLM2}} 
\newcommand{\llmthree}{\texttt{LLM3}} 
\newcommand{\llmfour}{\texttt{LLM4}} 
\newcommand{\llmfive}{\texttt{LLM5}} 

\definecolor{ronecolor}{rgb}{0.937, 0.745, 0.173}
\newtcbox{\reqbox}{on line,
  colframe=black,        
  colback=ronecolor,
  coltext=white,        
  boxrule=0.2pt,        
  arc=1.5pt,              
  boxsep=0.1pt,
  left=2pt,right=2pt,top=1pt,bottom=1pt,
}

\definecolor{codecolor}{rgb}{0.518, 0.671, 0.314}
\newtcbox{\codellmbox}{on line,
  colframe=black,        
  colback=codecolor,
  coltext=white,        
  boxrule=0.2pt,        
  arc=1.5pt,              
  boxsep=0.1pt,
  left=2pt,right=2pt,top=1pt,bottom=1pt,
}

\definecolor{analyzecolor}{rgb}{0.8549, 0.4667, 0.2588}
\newtcbox{\analyzellmbox}{on line,
  colframe=black,        
  colback=analyzecolor,
  coltext=white,        
  boxrule=0.2pt,        
  arc=1.5pt,              
  boxsep=0.1pt,
  left=2pt,right=2pt,top=1pt,bottom=1pt,
}

\definecolor{sumcolor}{rgb}{0.2824, 0.6118, 0.8157}
\newtcbox{\sumllmbox}{on line,
  colframe=black,        
  colback=codecolor,
  coltext=white,        
  boxrule=0.2pt,        
  arc=1.5pt,              
  boxsep=0.1pt,
  left=2pt,right=2pt,top=1pt,bottom=1pt,
}

\definecolor{tomato}{rgb}{1.0, 0.39, 0.28}
\definecolor{limegreen}{rgb}{0.196, 0.804, 0.196}
\definecolor{lightlimegreen}{rgb}{0.68, 1.0, 0.68}  
\definecolor{lighttomato}{rgb}{1.0, 0.7, 0.6}      
\newtcbox{\simcircle}{on line,
  colframe=black,        
  colback=lightlimegreen,     
  coltext=black,         
  boxrule=0.2pt,         
  arc=4pt,               
  boxsep=0.1pt,
  left=1.85pt,right=1.85pt,top=1pt,bottom=1pt,
}

\newtcbox{\diffcircle}{on line,
  colframe=black,        
  colback=lighttomato,     
  coltext=black,         
  boxrule=0.2pt,         
  arc=4pt,               
  boxsep=0.1pt,
  left=1.85pt,right=1.85pt,top=1pt,bottom=1pt,
}

\definecolor{codellmgreen}{rgb}{0.259, 0.686, 0.241} 

\newcommand{\codellm}{\textcolor{codellmgreen}
{\textbf{coding-LLM}}} 
\newcommand{\analyzellm}{\textcolor{analyzecolor}{\textbf{analysis-LLM}}}

\begin{document}


\firstsection{Introduction}

\maketitle

\label{sec:introduction}
Powerful LLMs have fostered the emergence of autonomous coding agents, which are capable of generating code, iteratively debugging/improving it, and ultimately producing functional code that rivals human-level expertise. The iterative solution-seeking process usually follows two major paradigms~\cite{yao2023tree}, chain-based and tree-based. The former commits to a chosen path and keeps iterating forward along that path for improvement. The latter branches into multiple paths and follows 
a certain policy to switch among them to explore a larger solution space. This work focuses on tree-based approaches as they are better suited for solution-seeking in complex coding tasks, e.g., building a machine learning (ML) model.
Within this realm, a recent coding framework, AI-driven exploration (AIDE)~\cite{aide,jiang2025aide}, has distinguished itself from others (e.g., LangChain, AutoML, and human-assisted ChatGPT) by excelling at coding challenges frequently encountered by ML scientists~\cite{wijk2024re,chan2024mle}. This work focuses on AIDE as a concrete case to propose a visual analytics system, but the system is not limited to AIDE and we also explore how it can be generalized to other coding frameworks.

AIDE operates through an iterative cycle of solution refinement. It first generates multiple code drafts, executes them using a compiler (Python interpreter), and analyzes the output. If the code runs successfully, it returns a target metric value (e.g., accuracy for classifications); if the code fails, it provides error messages. Based on this feedback, the agent autonomously modifies the code. This cycle continues until a termination criterion, such as a maximum iteration count, is met. While this process can be completely autonomous, it presents a major challenge: \textit{ML scientists have limited visibility into how the agent explores solutions and arrives at its final result}. Understanding this journey is crucial—not only for assessing the agent’s decision-making process but also for refining its behavior by adjusting prompts or selecting LLMs.

\setlength{\belowcaptionskip}{-10pt}
\begin{figure}[tb]%
  \centering 
  \includegraphics[width=\columnwidth]{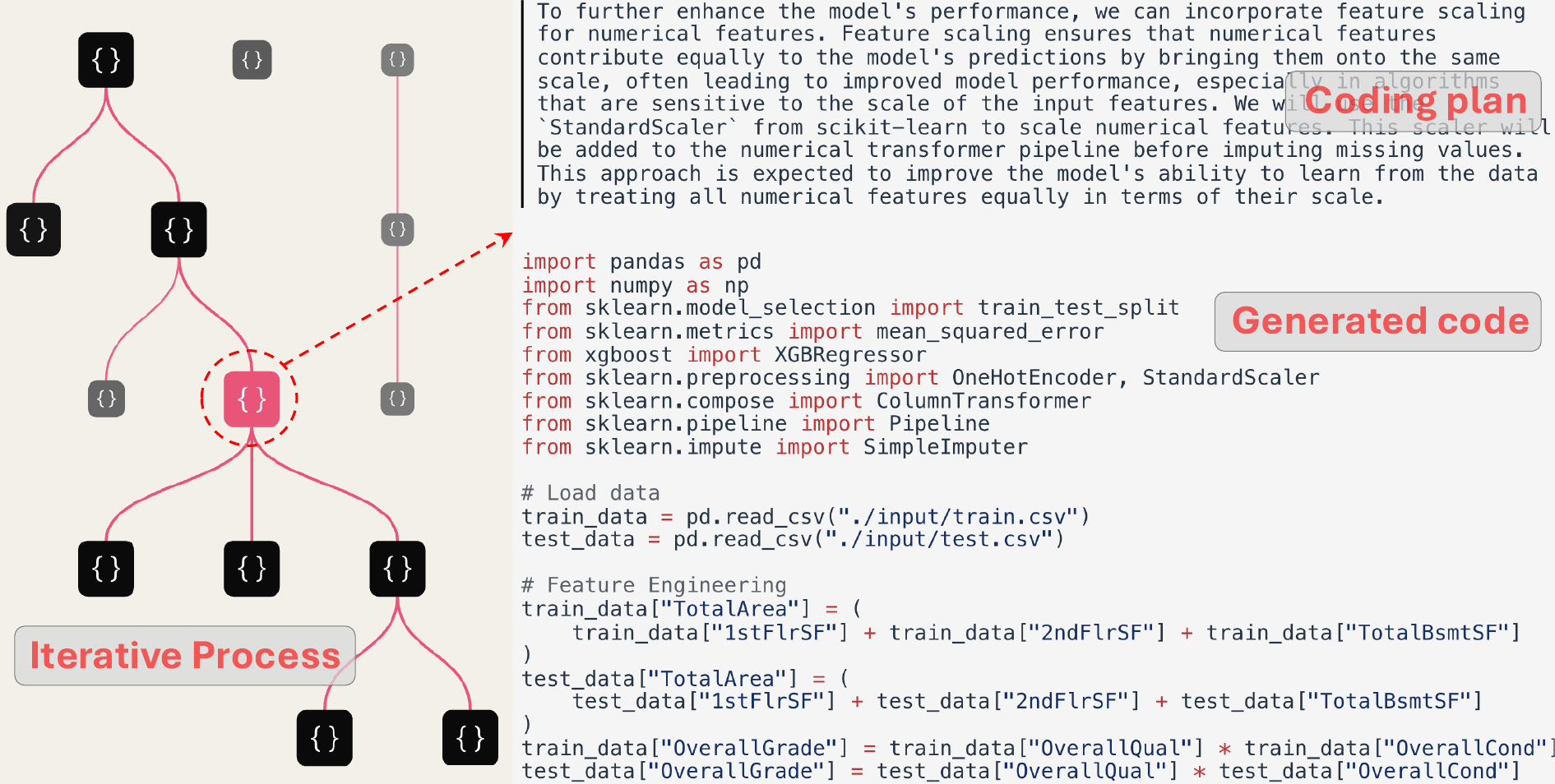}
  \vspace{-0.2in}
  \caption{The visualization tool from AIDE~\cite{aide, jiang2025aide}. (Left) The agent starts with a code \texttt{draft} (a root) and keeps \texttt{improving} or \texttt{debugging} it (extending the root to become a tree, and there are three trees in this view). (Right) The generated plan and code from the selected tree node.
  }
  \label{fig:aidevis}
\end{figure}
\setlength{\belowcaptionskip}{0pt}
To address this challenge, AIDE includes a tree-based visualization (Fig.~\ref{fig:aidevis}, left), which maps out the agent’s iterative solution-seeking process. Each tree starts with an initial code draft as its root, with subsequent nodes representing refinements or bug fixes. Clicking on a node reveals the corresponding code (Fig.~\ref{fig:aidevis}, right). While this visualization provides a high-level view of the agent’s coding trajectory, it lacks critical details that ML scientists need for deeper analysis. For instance, our collaborating experts found it difficult to identify:
\begin{enumerate}[leftmargin=0.7cm, labelsep=0.1cm, itemsep=-0.02cm]
    \item Whether the code included in a tree node is buggy or functional.
    \item How well (e.g.,  accuracy or loss) each solution performs.
    \item Which node represents the best-performing solution.
    \item How long each node takes to execute (code execution time).
    \item How different a node's code is compared to its parent.
    \item How the agent iteratively fixed a particular bug.
\end{enumerate}

Beyond tracking coding iterations within a single agent run, domain experts also seek deeper insights across different runs and LLMs. These insights can significantly improve their efficiency in refining the agent. However, they currently lack effective tools to obtain such insights.

To fill this gap, we introduce a visual analytics system that facilitates understanding of the agent’s coding process. The system not only enriches the existing visualization, but also introduces a structured three-level comparative framework for systematic agent analysis:
\begin{itemize}[leftmargin=0.55cm, labelsep=0.1cm, itemsep=-0.02cm]
    \item \textbf{{Code-Level}}: Highlights differences between two code versions, pinpointing how the agent debugs and improves a solution.
    \item \textbf{{Process-Level}}: Compares multiple solution-trees generated by the same LLM backbone, examining solution-seeking policies, execution time, and solution quality.
    \item \textbf{{LLM-Level}}: Evaluates coding behaviors across different LLMs, providing insights into their coding preferences.
\end{itemize}

To demonstrate the effectiveness of our system, we applied it to analyze AIDE’s performance on multiple Kaggle challenges. By dissecting the agent’s coding behavior, we uncovered insights into how different LLMs approach problem-solving, the strategies they employ, and how they refine code over iterations. In short, our contributions are twofold:
\begin{itemize}[leftmargin=0.55cm, labelsep=0.1cm, itemsep=-0.02cm]
    \item We designed and developed a visual analytics system that enhances transparency in coding agents' solution-seeking processes.
    \item We introduced a three-level comparative analysis framework, providing actionable insights into LLM-driven agentic coding.
\end{itemize}

\section{Related Work}
{\textbf{AI and Visualization.}}
The intersection of AI and visualization has given rise to two primary research directions. The first explores how AI can enhance traditional visualization algorithms (AI4VIS~\cite{wang2021survey}), such as graph drawing~\cite{kwon2019deep, wang2019deepdrawing} and volume rendering~\cite{berger2018generative, shi2019cnns}. The second investigates how visualization can be leveraged to better understand AI models (VIS4AI~\cite{hohman2018visual, wang2024visual, Liu2025}), e.g. making them more interpretable~\cite{liu2016towards, wang2018ganviz, li2023does, rathore2024verb,strobelt2017lstmvis}, diagnosing their interal issues~\cite{wexler2019if, wang2019deepvid, strobelt2018s, li2023visual}, steering their behavior~\cite{ming2019protosteer, li2020cnnpruner, yang2020interactive, strobelt2021genni}, and enhancing their performance~\cite{bilal2017convolutional, zhang2022sliceteller}. 

Our work is closely related to two topics within VIS4AI. The first is \textbf{\textit{agent visualizations}}, where most existing works have centered on using visualization to interpret agents trained from reinforcement learning (RL) models. For example, DRLViz~\cite{jaunet2020drlviz} and DynamicsExplorer~\cite{he2020dynamicsexplorer} use dynamic heatmaps to expose the memory patterns of RL agents and diagnose their decision-makings. DQNViz~\cite{wang2018dqnviz} and DRLive~\cite{wang2021visual} employ time-series data visualizations to examine the evolution of an agent' intelligence over time. 
\textit{In contrast, our work focuses on coding agents, especially the agents' iterative code debugging and improvement process, rather than their interaction with a specific RL environment.} 
The second is \textit{\textbf{comparative visual analysis of LLMs}}. For example, LLM Comparator~\cite{kahng2024llm, kahng2024llm1} puts LLMs into side-by-side visualization views to compare them and answer in what scenario one LLM outperforms the other.
ChainForge~\cite{arawjo2024chainforge} provides a friendly interface to help users compare LLM prompts and their responses in a table/list view, facilitating users to efficiently perform hypothesis testing.
EvalLM~\cite{kim2024evallm} allows users to define their own evaluation criteria and use LLMs to evaluate LLMs' outputs based on those criteria for better prompt engineering.
\textit{Different from these works, we focus on contrasting the code generated by LLMs and their coding preference.} 
The joint of the above two topics has resulted in many interactive visualizations for LLM-powered agentic systems. Dhanoa et al.~\cite{dhanoa2025agentic} have summarized them in their recent brief survey on agentic visualization. According to their survey and our knowledge till the written of this paper, however, \textit{visualizations for agentic coding systems have barely been covered in the visualization field, despite their demanding need.}

Apart from comparing agents and their backbone LLMs, it is also crucial to compare different agent runs of the same LLM. These runs are generated to measure the inherently non-deterministic nature of LLMs. The underlying visualization challenge closely parallels \textit{\textbf{ensemble visualization}}~\cite{obermaier2014future, wang2018visualization}, where a variety of techniques have been developed over the past decades, including side-by-side views, statistical summaries, and glyph-based representations~\cite{he2020cecav, he2019insitunet, sanyal2010noodles, potter2009ensemble}. \textit{We employ glyphs~\cite{ropinski2011survey} to encode the variations across different solution-seeking processes and apply clustering to facilitate their exploration.} 

{\textbf{Solution-Seeking in Agentic Coding.}}
LLM-assisted agentic coding scaffolds~\cite{githubcopilot, geminiassist, wang2021codet5, le2022coderl, wang2023codet5plus} typically involve iterative interactions between a code generator (the LLM agent) and a code interpreter to converge on optimal solutions. Two principal paradigms have emerged for exploring the solution space~\cite{liu2025ml}.
The first is the \textbf{\textit{chain-based}} approach, exemplified by OpenHands~\cite{wang2024openhands}, LangChain~\cite{langchain2023} and CodeChain~\cite{le2023codechain}, which modularize individual components and compose them into sequential workflows. This chain structure enables clear, stepwise execution and facilitates straightforward interpretation of the logic flow.
The second paradigm is the \textit{\textbf{tree-based}} exploration, where a tree structure is constructed with multiple branches, each representing a distinct direction of solution search. Notable examples include AlphaZero~\cite{wan2024alphazero} and scattered forest search~\cite{lightsfs}, both of which leverage Monte Carlo tree search to balance exploration and exploitation across the solution space. Recent research has increasingly focused on this paradigm, investigating how the width and depth of the search tree affect solution quality and efficiency~\cite{misaki2025wider}.
Among tree-based coding agents, AIDE~\cite{aide, jiang2025aide} has stood out from its counterparts due to its superior performance in \textit{\textbf{solving ML problems}} (the focused coding problems in this work), according to the latest benchmarks~\cite{chan2024mle, wijk2024re}.
Given its effectiveness, our work centers on AIDE, aiming to visualize its solution-seeking process and conduct comparative analyses at multiple levels. \textit{While we focused on AIDE to concretize our visual designs in this work, we strive to make our designs general-purpose and discuss how our proposed system can be generalized to other tree-based coding frameworks in Sec.~\ref{sec:discussion}.}

\section{AI-Driven Exploration (AIDE)}
\label{sec:background}
\begin{figure}[tb]
  \centering
  \includegraphics[width=\columnwidth, alt={A line graph showing paper counts between 0 and 160 from 1990 to 2016 for 9 venues.}]{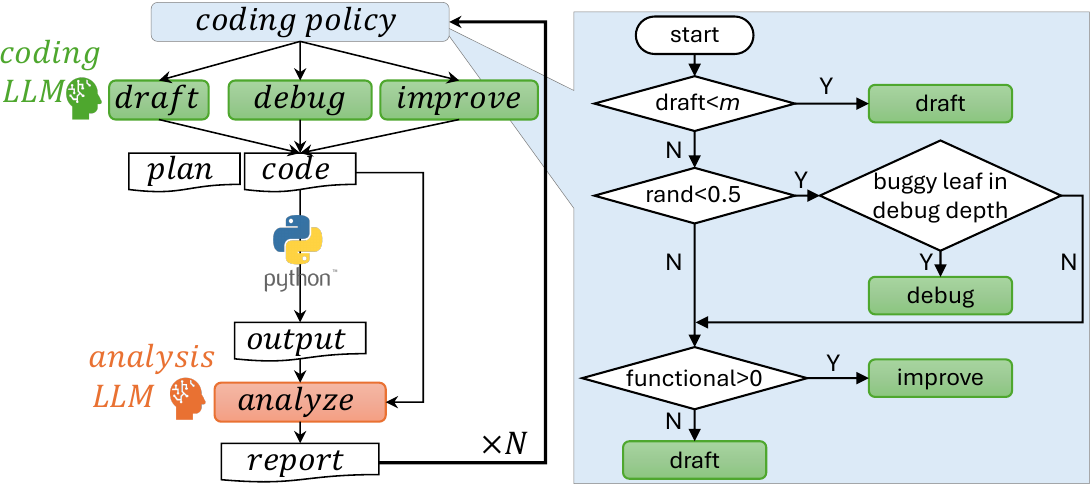}
  \vspace{-0.2in}
  \caption{The AIDE framework (left) and its coding policy (right).}
  \label{fig:strategy}
\end{figure}

Given an ML problem, the AIDE agent is directed to generate code and progressively refine it to optimize a target evaluation metric. This iterative solution-seeking process involves two key components (Fig.~\ref{fig:strategy}): (I) the LLM backbone and (II) the coding policy.

The first component consists of two LLMs that operate in alternation:
\begin{enumerate}[labelsep=0.1cm, itemsep=-0.02cm]
    \item The \codellm{} takes (1) the problem and data description, (2) the evaluation metric, and (3) the feedback from earlier iterations as input to generate (1) a paragraph outlining its coding \underline{plan} and (2) a piece of \underline{code} based on the plan. Then, the agent calls Python to execute the code and saves its \underline{output}, which may be error logs for \textit{buggy} code or a metric value for \textit{functional} code.

    \item The \analyzellm{} takes (1) the coding plan, (2) the generated code, and (3) the execution output as input to evaluate the current iteration and generate an analysis \underline{report} as feedback. 
\end{enumerate}

\noindent The two LLMs are repeatedly called for $N$ iterations to generate $N$ solutions, organized as a forest of trees (Fig.~\ref{fig:aidevis}, left). This work focuses on the coding-LLM, as it is the key for code generation. We use five different LLM backbones (\llmone{}-\llmfive{}\footnote{All are the latest LLMs as of early 2025. We hide their names to ensure LLM agnosticism and prevent user bias toward different models in evaluations.}) and compare their coding behaviors. For the analysis-LLM, we consistently use the same one.

The second key component decides if the agent should generate code from scratch (\texttt{Draft}), fix a bug from an earlier iteration (\texttt{Debug}), or improve previously functional code (\texttt{Improve}). The flowchart in Fig.~\ref{fig:strategy} illustrates this policy. 
At the start of each iteration, the agent checks the current number of iterations and terminates the process if $N$ ($N{=}30$), the termination criterion, has reached.
Otherwise, the agent follows these steps to generate additional solutions/nodes:

\begin{enumerate}[leftmargin=0.6cm, labelsep=0.1cm, itemsep=0.01cm]
    \item If the number of \texttt{draft} nodes is less than $m$ ($m{=}5$ in this work), the agent continues proposing new \texttt{drafts}. 
    \item If there are $m$ or more \texttt{drafts}, the agent randomly decides whether to pick a buggy node to \texttt{debug} or a functional node to \texttt{improve}. 
    
    \item When \texttt{debugging}, the agent randomly picks a buggy leaf node that is within the specified tree depth to debug. If no node satisfies the criteria, go to the \texttt{improve} operation.
    
    \item If the agent needs to \texttt{improve} a functional node but all existing nodes are buggy, it proposes a new \texttt{draft}. Otherwise, it selects the \textbf{\textit{best functional node}} to \texttt{improve}, following a greedy heuristic. 
\end{enumerate}

The three operations share the same coding-LLM, but use distinct prompts~\cite{aideprompt} tailored to their respective goals.
The iterative calling of the three operations results in a forest of trees (Fig.~\ref{fig:aidevis} shows three trees). Each \texttt{draft} node serves as the root of a tree, with \texttt{debug} and \texttt{improve} nodes expanding the tree. Here is a summary of our terminology:
\begin{itemize}
    \item \textit{\textbf{Solution}}: A single tree node containing the coding plan and generated code. The node can be a \texttt{draft}, \texttt{debug}, or \texttt{improve} node. It is considered \texttt{functional} if the code executes successfully and produces an evaluation metric value, or \texttt{buggy} if not. 
    
    \item \textbf{\textit{Solution-Seeking Process}}: $N$ tree nodes ($N{=}30$ in this work), organized as a forest of trees, generated in one complete run of AIDE. All trees in the forest will later be merged into a larger tree, referred to as a \textit{\textbf{Solution-Tree}}.
\end{itemize}

\setlength{\belowcaptionskip}{-15pt}
\setlength{\belowcaptionskip}{0pt}

Visualizing the code generation process is crucial to monitor the agent's behavior.
The developers of AIDE recognized this and introduced a visualization tool alongside their work. 
As shown in Fig.~\ref{fig:aidevis}, the tool provides two views. On the left, node-link diagrams illustrate the iterative process of \texttt{drafting} a solution (represented by the three tree roots) and \texttt{debugging}/\texttt{improving} the drafts to expand the root into a tree. When a node is selected, its coding plan and generated code will be shown on the right.
As mentioned in the Introduction, ML scientists have identified several limitations of the tool.
More importantly, it falls short of meeting the scientists' needs for performing in-depth analyses and comparisons of the generated code, solution-seeking processes, and LLMs, which motivates the development of our system.
\setlength{\belowcaptionskip}{-10pt}
\textbf{\begin{figure*}[tb]
  \centering 
  \includegraphics[width=\textwidth, alt={A line graph showing paper counts between 0 and 160 from 1990 to 2016 for 9 venues.}]{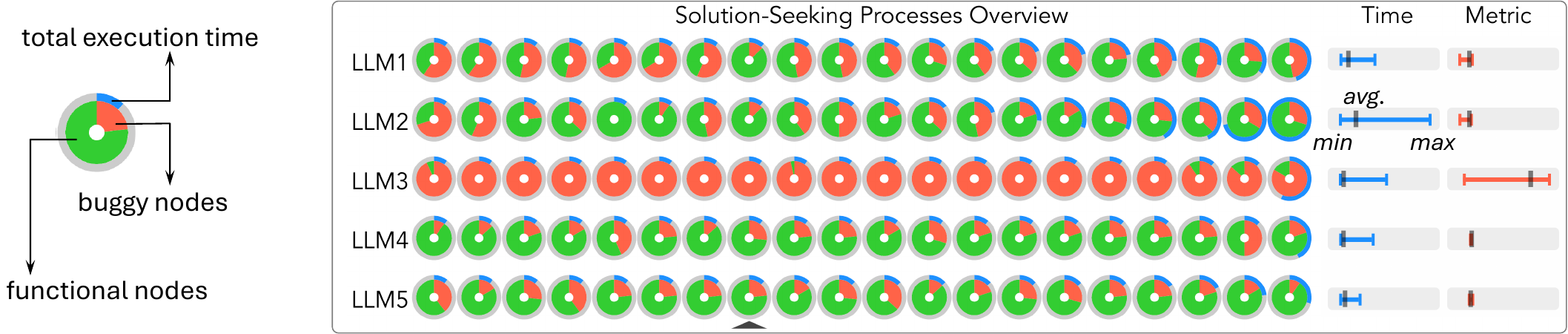}
  \vspace{-0.2in}
  \caption{Comparing 100 solution-trees from 5 LLMs. The pie charts of each row are ordered by the total execution time (see the outer blue arcs).}
  \label{fig:root}
\end{figure*}}
\setlength{\belowcaptionskip}{0pt}

\section{Requirement Analysis}
Our target users are \textit{ML scientists who design and develop coding agents}, rather than the end users who simply apply them. Our aim is to surface nuanced insights into agent behavior to help the scientists better understand and enhance the agents they build. To ground our work in real-world needs, we partnered with five ML scientists from a research lab. Each scientist holds a Ph.D. in computer science and brings over four years of hands-on industry experience building production-scale ML systems. Right now, their focus is sharp: building Python-based coding agents that empower internal ML teams to develop better ML models. Agents for other coding tasks or those using other programming languages are outside their focus. While they have worked with multiple coding frameworks, AIDE~\cite{aide} remains their primary focus.

We worked intensively with two of the five experts to capture both their own and their colleagues’ experiences using AIDE and its visualization tool. The remaining three experts joined later as objective participants in case studies to assess our visualization system. The two primary experts pointed out several limitations of the visualization tool in Fig.~\ref{fig:aidevis} (see Introduction). For example, after each run, they wanted immediate answers to key questions: What was the best performance achieved, and in which iteration? What were the major bugs, and how did the agent attempt to fix them? While the AIDE visualization tool helps trace the iterative coding process, its analytical power is limited, given that it wasn’t designed for in-depth analysis. The experts also emphasized the need for comparing code and agent runs, which they believed would significantly boost their working efficiency. Through iterative refinement with them, we have distilled three core requirements that form the foundation of a comparative analysis framework.

\begin{itemize}[topsep=0pt, partopsep=0pt, leftmargin=0.5cm, labelsep=0.1cm, itemsep=-0.02cm]
\item \textbf{\reqbox{R1} Code-Level Analysis}: Analyzing the generated code requires access to all relevant contextual information. For example: What solution does the code implement? Does it encounter any bugs? If so, what are they? If not, how well does it perform? 
Additionally, comparing two versions of code is often necessary to understand how the agent fixes bugs or improves previous implementations. These needs translate into the following system requirements:
\begin{itemize}[topsep=0pt, partopsep=0pt, leftmargin=0.5cm, labelsep=0.1cm, itemsep=-0.02cm]
    \item \textbf{R1.1} - Present supporting information for the generated code, including the coding plan, the code itself, execution logs, performance metrics, runtime, bug status, execution reports, etc.
    \item \textbf{R1.2} - Compare two code, especially the code between consecutive iterations, to analyze bug fixes or code improvements.
\end{itemize}

\item \textbf{\reqbox{R2}} \textbf{Process-Level Analysis:} Understanding the agent's behavior requires analyzing how it creates new \texttt{drafts} from scratch and how it \texttt{debugs} or \texttt{improves} existing ones within a solution-seeking process. Thus, visualizing the complete process and the relationships between individual nodes is essential. 
Moreover, since the agent’s behavior is inherently non-deterministic, the same process often needs to be run multiple times. Comparing these runs helps quantify uncertainty and better assess the agent's consistency in its coding policy. To support these, our system needs to:
\begin{itemize}[topsep=0pt, partopsep=0pt,leftmargin=0.5cm, labelsep=0.1cm, itemsep=-0.02cm]
    \item \textbf{R2.1} - Present a solution-seeking process with clearly annotation of the relationships between iterations, and the statistics of the process, such as total running time, buggy node ratio, etc.
    \item \textbf{R2.2} - Compare multiple processes to assess how similar they are and to analyze the agent’s coding policy across runs.
\end{itemize}
    
\item \textbf{\reqbox{R3} LLM-Level Analysis}: One question that often puzzles our experts is whether different coding-LLMs behave noticeably differently—and if so, what sets them apart. When it comes to the generated ML solutions, the experts care deeply about each LLM’s model preferences (e.g., does one favor linear models, while another prefers tree-based models?), as well as their efficiency and overall performance. Gaining insight into these differences is crucial for making informed choices about which LLM to deploy for specific tasks. Therefore, our system needs to:
\begin{itemize}[topsep=0pt, partopsep=0pt,leftmargin=0.5cm, labelsep=0.1cm, itemsep=-0.02cm]
    \item \textbf{R3.1} - Compare the code generated by different LLMs to reveal their distinct solution styles and model/package preferences.
    \item \textbf{R3.2} - Highlight process-level differences across LLMs, showing which models produce fewer bugs or complete tasks faster.
\end{itemize}

\end{itemize}

\section{Visual Analytics System}
Focusing on the three requirements, we propose a structured three-level comparative analysis framework and implement it through a visual analytics system with four coordinated views (Figs.~\ref{fig:system1} and~\ref{fig:system2}).

\subsection{The \treeview{}}
The \treeview{} (Fig.~\ref{fig:system1}a) visualizes the solution-seeking process as a node-link diagram. This representation was chosen over other tree visualizations, e.g., treemaps and dendrograms, due to its intuitiveness to the experts.
A process contains $N{=}30$ nodes in at least $m{=}5$ trees. 
To ease the exploration, we merge trees in the same process into a single larger tree through an extra root, resulting in $N{+}1$ nodes in total. This root is visualized as a pie chart to indicate the distribution of functional and buggy nodes in the entire solution-seeking process. The following lists the visual encoding details in Fig.~\ref{fig:system1}a (\textbf{R1.1}, \textbf{R2.1}):

\begin{itemize}[topsep=2pt, partopsep=0.01pt, leftmargin=0.5cm, labelsep=0.1cm, itemsep=-2pt]
    \item Node color: green $\Rightarrow$ functional node;  red $\Rightarrow$ buggy node.
    \item Number inside a node: the step ID in the solution-seeking process. This ID is underlined if the node is an internal node.
    \item Numerical value next to a functional node: the evaluation metric.
    \item Award ribbon: the best-performing node (i.e., node 28 in Fig.~\ref{fig:system1}a).
    \item Outer blue-arc surrounding a node: code execution time, the fuller the arc the longer the time (e.g., node 17 has the longest time).
    \item Link thickness: the magnitude of code changes between a parent-child pair, measured by the number of modified lines of code.
\end{itemize}

Clicking on an internal node collapses the subtree from that node, making it easier to navigate a large tree (e.g., the trees in~\cite{chan2024mle, misaki2025wider} have hundreds of nodes). The underlined step ID of internal nodes serves as a visual cue, indicating which nodes can be expanded when collapsed.
Selecting a node displays its coding plan, generated code, execution logs, etc. In Fig.~\ref{fig:system1}b, the plan for node 11 reveals that it was generated to \texttt{debug} its parent, node 8. The plan identifies the issue in node 8—using \texttt{XGBRegressor}, which is unavailable in the environment—and outlines the agent’s fix by replacing it with \texttt{GradientBoostingRegressor}.
The corresponding code and analysis report are displayed in the \codeview{}, which will be explained later in Figs.~\ref{fig:system1}c and~\ref{fig:system1}d.

Visualizing the roots of different solution-trees allows us to efficiently compare different processes (\textbf{R2.2}).
Fig.~\ref{fig:root} presents 100 solution-seeking processes, each represented by the root of the corresponding tree. 
These roots are arranged into 5 rows. Each row is the result of using a unique coding-LLM to run the solution-seeking process 20 times.
To the right, two columns of glyphs display the aggregated statistics for execution time and evaluation metric per row. The start, end, and black ticks on each glyph mark the minimum, maximum, and average value.
Together, the root and glyph visualizations provide an overview of all solution-seeking processes, facilitating intuitive comparisons between different processes and LLMs (\textbf{R2.2}, \textbf{R3.2}). For instance, Fig.~\ref{fig:root} clearly shows that \llmthree{} tends to generate more buggy nodes, while \llmtwo{}'s code typically requires more execution time.

To quickly locate a tree for drill-down analysis, the root nodes within the same row can be ordered according to one of the following metrics:
\begin{itemize}[topsep=2pt, partopsep=0.01pt, leftmargin=0.5cm, labelsep=0.1cm, itemsep=-2pt]
    \item The \textit{\textbf{total time}} used to generate the entire tree, encoded by the filling ratio of the blue outer arc surrounding each pie chart (Fig.~\ref{fig:root}, left).
    \item The \textit{\textbf{best evaluation metric value}} out of all nodes of the tree.
    \item The \textit{\textbf{number of buggy/functional nodes}} in the entire tree.
    \item The \textit{\textbf{tree structure similarity}} based on tree-edit distance~\cite{zhang1989simple}. 
\end{itemize}

\begin{figure}[bh]
  \centering 
  \includegraphics[width=\columnwidth, alt={A line graph showing paper counts between 0 and 160 from 1990 to 2016 for 9 venues.}]{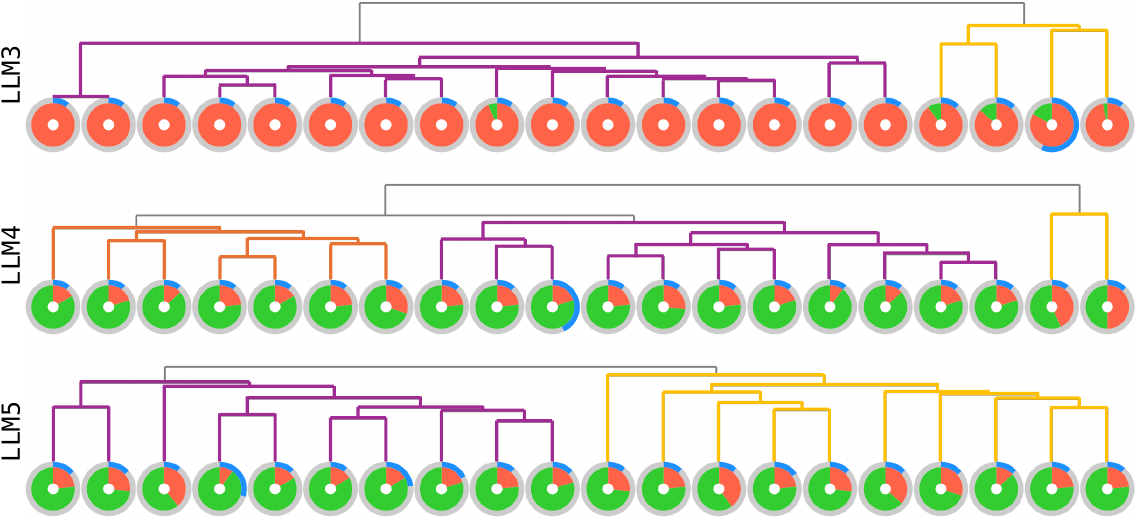}
\vspace{-0.2in}
\caption{Clustering and ordering the roots of solution-trees from the same LLM based on the tree-edit distance between the corresponding trees.}
  \label{fig:ted}
\end{figure}

For the tree structure similarity ordering, we use the tree-edit distance, which counts the number of edits (such as adding or removing nodes) required to transform one tree into another. We conduct pair-wise comparison over all $k$ ($k{=}20$) trees in a row to generate a $k{\times}k$ matrix. Using it, we hierarchically cluster the trees and display the result in a dendrogram.
Fig.~\ref{fig:ted} shows the dendrograms for \llmthree{}-\llmfive{}. The clustering helps guide users' exploration of the trees, as they only need to focus on one representative tree from each cluster. For example, most of the solution-trees from \llmthree{} are grouped into the purple cluster. These trees tend to be wide and contain mostly buggy nodes, as explained later in Sec.~\ref{sec:casestudy} and Fig.~\ref{fig:buggy}.

\setlength{\belowcaptionskip}{-10pt}
\setlength{\belowcaptionskip}{0pt}

\setlength{\belowcaptionskip}{-10pt}
\begin{figure}[tb]
  \centering 
  \includegraphics[width=\columnwidth, alt={A line graph showing paper counts between 0 and 160 from 1990 to 2016 for 9 venues.}]{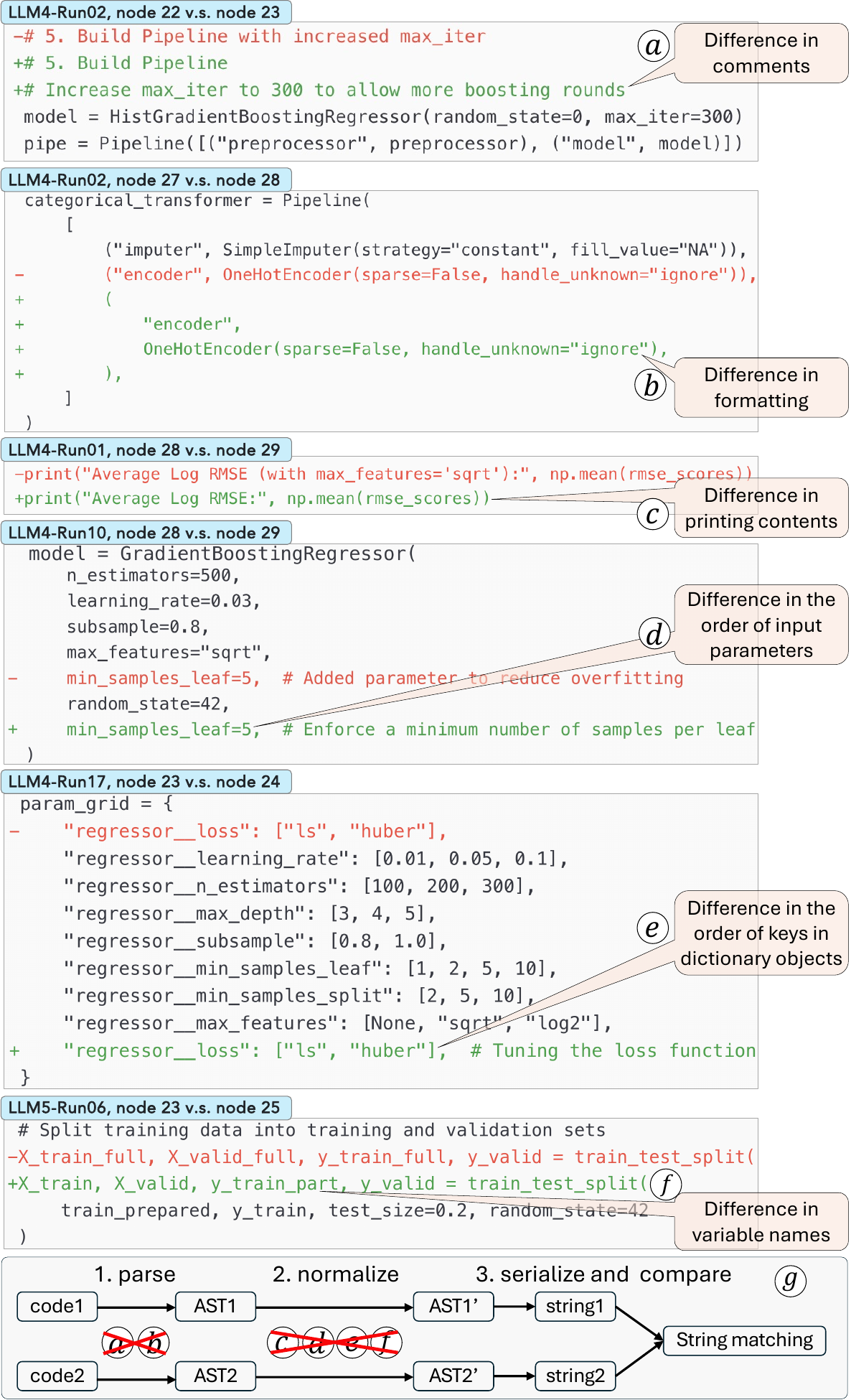}
  \vspace{-0.21in}
  \caption{Trivial code differences (a-f) and our pipeline to exclude them (g).}
  \label{fig:codediff}
\end{figure}
\setlength{\belowcaptionskip}{0pt}

\setlength{\belowcaptionskip}{-10pt}
\begin{figure}[tb]
  \centering 
  \includegraphics[width=\columnwidth, alt={A line graph showing paper counts between 0 and 160 from 1990 to 2016 for 9 venues.}]{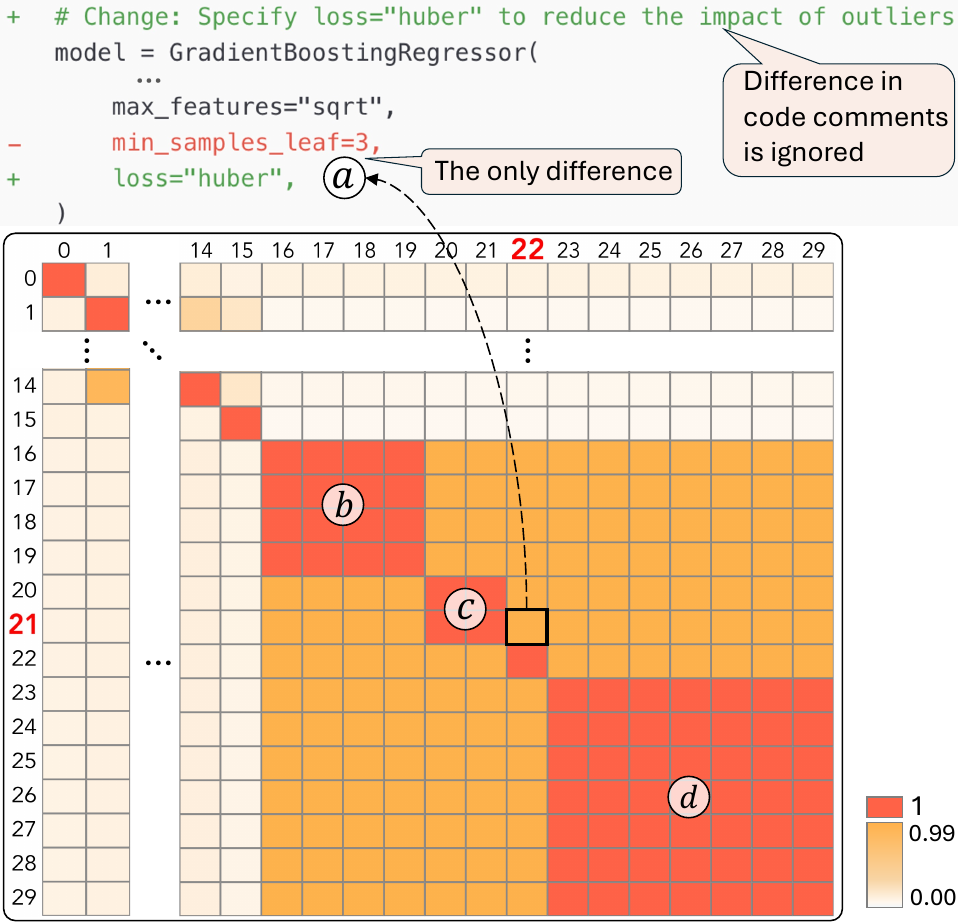}
  \vspace{-0.21in}
  \caption{Code similarity among the 30 nodes of the tree in Fig.~\ref{fig:twotree}a.}
  \label{fig:matrix}
\end{figure}
\setlength{\belowcaptionskip}{0pt}

\subsection{The \codeview{}}
The \codeview{} (Fig.~\ref{fig:system1}c) displays the code of the selected node, helping users examine it.
Beyond simply presenting the code, it is more important to reveal the subtle changes made during \texttt{debugging}/\texttt{improving}, i.e., comparing two versions of code (\textbf{R1.2}). Therefore, we introduce a ``\textit{Code Difference}'' mode. When enabled, the view highlights the \textbf{\textit{line-level}} differences between code in two consecutively selected tree nodes.
Fig.~\ref{fig:system1}c contrasts the code in nodes 8 and 11 (\textcolor{darkgray}{black: shared lines}; \textcolor{RedOrange}{-red: lines in node 8 only}; \textcolor{YellowGreen}{+green: lines in node 11 only}), and discloses how the agent replaces \texttt{XGBRegressor} with \texttt{GradientBoostingRegressor} to fix the bug in node 8. This explicit highlighting enables more efficient comparison than manually switching between nodes and checking their code line by line.

Besides, the experts noted that many code changes between iterations are purely cosmetic—such as tweaks in variable names or formatting styles. Fig.~\ref{fig:codediff}a-\ref{fig:codediff}f show six representative cases where the line-level difference leads to false code difference. Yet, what truly matters is whether these changes alter the code’s functionality.

To capture the real code difference, we introduce a \textbf{\textit{function-level}} code similarity score, which reaches 1 when two code versions function the same. Specifically, we strip away superficial changes and assess functional equivalence by comparing the code's abstract syntax trees (ASTs)~\cite{alfred2007compilers} in three steps (Fig.~\ref{fig:codediff}g):

\begin{enumerate}[topsep=2pt, partopsep=0.01pt, leftmargin=0.35cm, labelsep=0.1cm, itemsep=-1.5pt]
    \item Parse the two versions of code into two ASTs, which automatically excludes comments and formatting differences in Figs.~\ref{fig:codediff}a and~\ref{fig:codediff}b.
    \item Traverse both ASTs and perform the following normalizations:
    \begin{itemize}[topsep=-3pt, partopsep=0.01pt, leftmargin=0.2cm, labelsep=0.1cm, itemsep=-2pt]
        \item if the visited node of the AST is a function call and the function name is `print', prune that tree branch (for the case in Fig.~\ref{fig:codediff}c).

        \item for function calls, also sort their
        input arguments by keyword name (for the case in Fig.~\ref{fig:codediff}d).
        
        \item when visiting a \texttt{dictionary} object, sort the keys of it, since key order does not affect functionality  (for the case in Fig.~\ref{fig:codediff}e).

        \item consistently rename all identifiers such as variables (e.g., rename them to \texttt{var1},  \texttt{var2}, etc.), functions, and classes (Fig.~\ref{fig:codediff}f).
        
    \end{itemize}
   
    \item Serialize the two ASTs into plain strings, then compute the similarity score between them using a sequence-matching algorithm~\cite{ratcliff1988pattern}.
\end{enumerate}

We compare all pairs of code in the same solution-tree and visualize the resulting $N{\times}N$ similarity matrix as a heatmap. In Fig.~\ref{fig:matrix}, the similarity values from 0 to 0.99 are mapped to colors from white to orange, 1 is mapped to red to highlight nodes with functionally identical code. The heatmap clearly shows that code in nodes 16-29 are very similar (shown in the orange block). For example, the difference between code in nodes 21 and 22 is only one parameter of the \texttt{GradientBoostingRegressor} (Fig.~\ref{fig:matrix}a). Furthermore, code in nodes 16-19 (Fig.~\ref{fig:matrix}b), nodes 20 and 21 (Fig.~\ref{fig:matrix}c), and nodes 23-29 (Fig.~\ref{fig:matrix}d) are functionally identical. 
The corresponding solution-tree is shown later in Fig.~\ref{fig:twotree}a. From it, we notice a severe issue of the agent in unintentionally repeating the same code when \texttt{improving} nodes 12 and 22, wasting computational resources.

At the bottom of the \codeview{}, the analysis report from the analysis-LLM is displayed as a paragraph of text (Fig.~\ref{fig:system1}d), evaluating and summarizing the agent's behavior in the selected node. Other information, such as the code execution logs, can also be shown here on-demand.

\setlength{\belowcaptionskip}{-10pt}
\begin{figure*}[tb]
  \centering 
  \includegraphics[width=\textwidth, alt={A line graph showing paper counts between 0 and 160 from 1990 to 2016 for 9 venues.}]{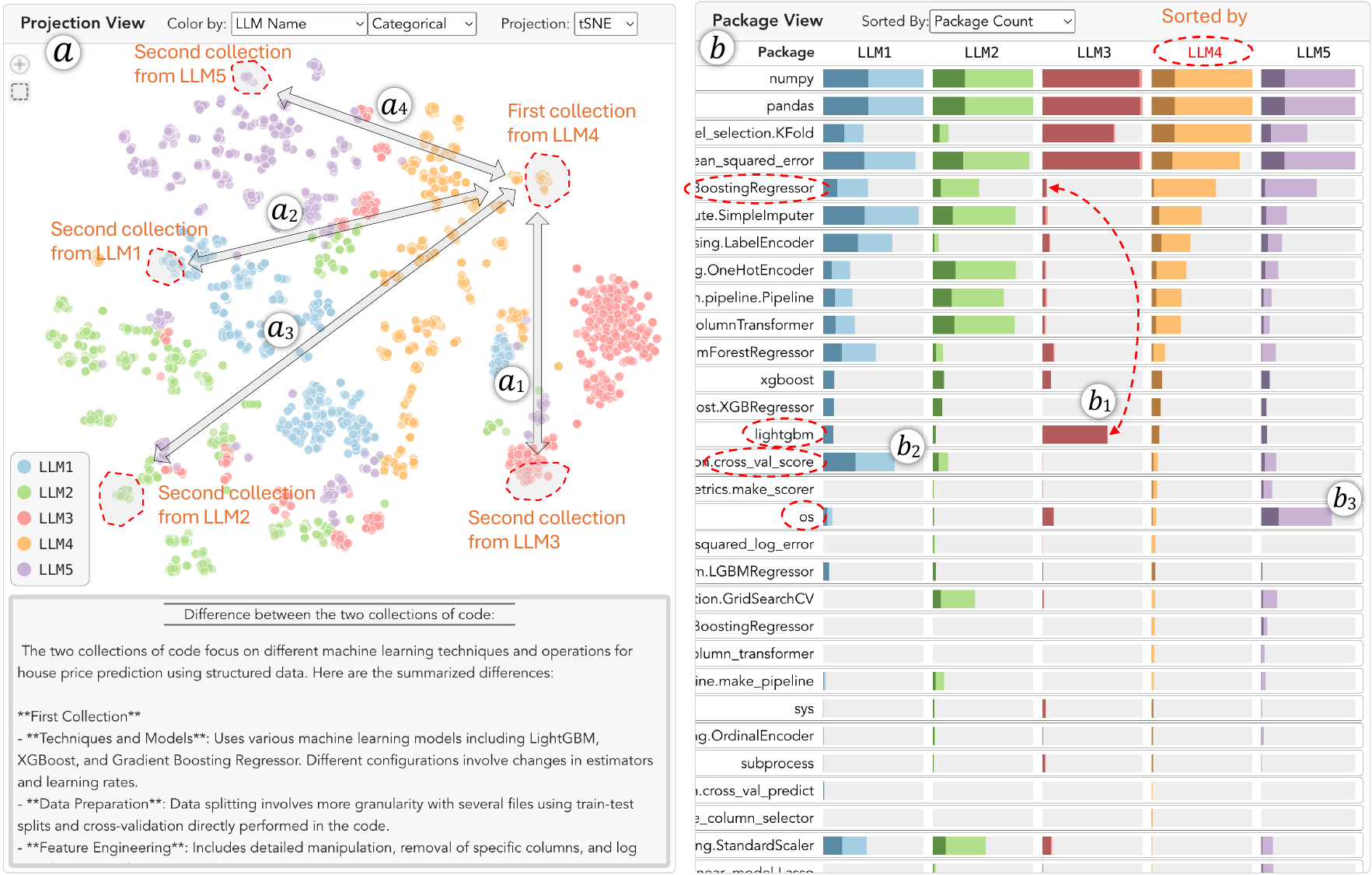}
  \vspace{-0.22in}
  \caption{Comparing the code generated by different LLMs. (a) The \projectionview{} embeds code into a high-dimensional space and uses dimensionality reduction (DR) algorithms to cluster them. (b) The \packageview{} conducts AST analysis~\cite{alfred2007compilers} on the code to count the frequency of different packages.}
  \label{fig:system2}
\end{figure*}
\setlength{\belowcaptionskip}{0pt}
\subsection{The \projectionview{}}
One key aspect that our experts always wanted to investigate on is the difference between code generated by different coding-LLMs (\textbf{R3.1}).
The \projectionview{} (Fig.~\ref{fig:system2}a) is designed for this purpose. It embeds each piece of code as a high-dimensional (300D) vector using a text-embedding model, and projects these vectors for different code to 2D for comparisons.
In Fig.~\ref{fig:root}, we have 100 solution-trees, each containing $N{=}30$ nodes, resulting in a total of 3,000 pieces of code. They are visualized as 3,000 points in the scatterplot in Fig.~\ref{fig:system2}a. Each point is colored by the used coding-LLM. From the point distribution, we can observe clear cluster separations by point color, indicating that different coding-LLMs exhibit dissimilar coding behaviors. Three popular DR algorithms, PCA, tSNE~\cite{van2008visualizing}, and UMAP~\cite{mcinnes2018umap}, are equipped in this view and users can switch among them to explore different layouts.
Additionally, zooming and lasso selection are both enabled in this view, allowing users to flexibly explore any large set of points.

The cluster pattern in Fig.~\ref{fig:system2}a confirms the existence of differences between coding-LLMs, but it does not tell what the differences are. To answer this, we leverage the language capability of LLMs for an  explicit summary. Specifically, users can select any two clusters of points from the scatterplot via lasso selections. The two collections of code are then concatenated respectively to form two strings, \textit{code1} and \textit{code2}. 
We then prompt a separate LLM with the following question:

\textit{``You are given two collections of code. Summarize the difference between them. The first collection is \{code1\}, the second collection is \{code2\}. Please be concise in your response and use bullet points.''}

The output from the separate LLM is displayed below the scatterplot. In Fig.~\ref{fig:system2}-a1, when comparing \llmfour{} (\textit{code1}) and \llmthree{} (\textit{code2}), the output is as follows (the full details are available in our Appendix):

\textit{``... Overall, the first collection presents a more traditional approach using multiple libraries and manual configurations, while the second leverages more of LightGBM's built-in functionalities for efficient data handling and model training."}

From the description, the main issue for \llmthree{} is that it is very biased towards \texttt{lightgbm}. This inflexibility answers why the code generated by \llmthree{} is often buggy (most pie portions are in red in Fig.~\ref{fig:root}), as the \texttt{lightgbm} package is not available in our Python environment.

\subsection{The \packageview{}}
Beyond the overview offered by the \projectionview{}, the experts also seek concrete, code-level evidence to support the differences in LLMs' coding behaviors. 
As the behavior of Python code largely depends on the packages it imports, the \packageview{} enables the comparison of LLMs by analyzing the frequency of their package usage (\textbf{R1.2}, \textbf{R3.1}).

In our setting, each coding-LLM generates 20 solution-trees and each tree contains 30 nodes, resulting in 600 code snippets per LLM and 3,000 in total across 5 LLMs.
Using AST~\cite{alfred2007compilers} analysis, we first identify the unique packages used across all 3,000 code snippets, and then, count their occurrences within each LLM's code. 
The resulting package frequencies are visualized in the \packageview{} (Fig.~\ref{fig:system2}b) as a matrix of bar charts. Each row represents a unique package, and each column represents a coding-LLM. The length of each bar represents the number of times that the corresponding package has been used by that coding-LLM. The bar's color matches the LLM's color in the \projectionview{}. Additionally, the dark-shaded portion of a bar reflects the ratio of buggy code (buggy/total nodes) when using that package. A ratio of 1 implies that the code is always buggy when the package is used, suggesting it is likely the root cause of a bug.

Clicking the name of each coding-LLM in the title row sorts the packages by their count decreasingly. In Fig.~\ref{fig:system2}b, the rows are sorted based on the package count from \llmfour{}. By comparing package counts across columns, we can identify notable behavioral differences across the coding-LLMs. For example, \llmthree{} uses \texttt{lightgbm} (Fig.~\ref{fig:system2}-b1) far more frequently, while others prefer \texttt{GradientBoostingRegressor}. This aligns with the findings from the \projectionview{}, explaining why the code generated by \llmthree{} is often buggy. The bar for \texttt{lightgbm} is also heavily shaded, indicating it may be the root cause of these bugs. Other coding-LLMs have recognized the unavailability of this package and substituted it with \texttt{GradientBoostingRegressor}.

\section{Case Study, Actionable Insights, and Feedback}
\label{sec:casestudy}
Together with the five ML scientists introduced in Sec.~\ref{sec:background}, we conducted case studies using our system to analyze the behavior of AIDE on 24 Kaggle competitions, 22 from the MLE-Bench (lite)~\cite{chan2024mle} and 2 from the AIDE code repository~\cite{aidecode}. 
Rather than detailing all competitions, this section focuses on a single competition, the {House-Price} prediction~\cite{houseprice}, to showcase how our system is used and the unique insights it can uncover. 
We selected this competition because it is included in the example cases of AIDE, allowing readers to easily reproduce our results.
Explorations of other competitions are included in our Appendix.

The {House-Price}~\cite{houseprice} dataset is a tabular dataset. Each row represents a house with features like \textit{Lot Area}, \textit{House Style}, \textit{Year Built}, \textit{Roof Style}, and \textit{Sale Price}. We prompt the agent to generate code of an ML model using the following problem description and evaluation metric:

\noindent\textbf{Problem}: \textit{Predict the sales price for each house.}

\noindent\textbf{Evaluation}: \textit{Use the RMSE metric between the logarithm of the predicted and observed values.}

Five coding-LLMs (\llmone{}–\llmfive{}) were used and each LLM repeated the solution-seeking processes for $k{=}20$ times. As a result, we generated 100 solution-trees. Inside each tree, the number of solution-seeking steps (tree nodes) is $N{=}30$, and the number of initial drafts is $m{=}5$. 

\setlength{\belowcaptionskip}{-10pt}
\begin{figure*}[tb]
  \centering \includegraphics[width=\textwidth, alt={A line graph showing paper counts between 0 and 160 from 1990 to 2016 for 9 venues.}]{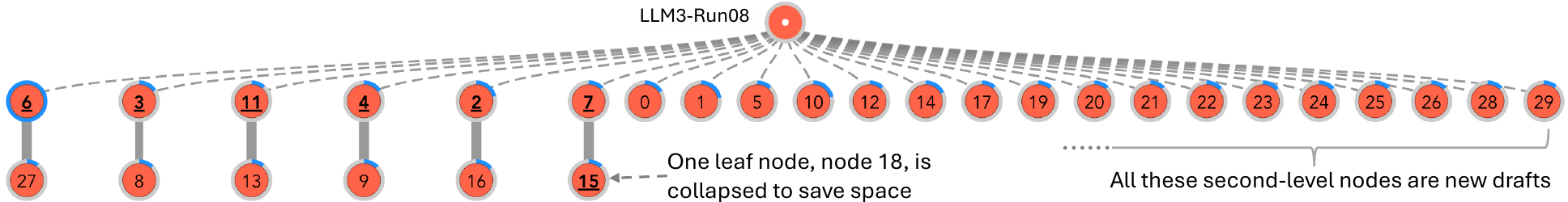}
  \vspace{-0.2in}
  \caption{A representative tree from \llmthree{}. It is very wide as there are no or very few functional nodes and the agent keeps generating new \texttt{drafts}.}
  \label{fig:buggy}
\end{figure*}
\setlength{\belowcaptionskip}{0pt}

\subsection{Comparative Exploration}
This section shows how our system supports users' comparative exploration across the three proposed levels. Although conceptually distinct, the three levels are often intertwined in practice. As such, each of the following subsections covers two levels at once.

\subsubsection{LLM-Level and Process-Level Comparison}
Fig.~\ref{fig:root} presents an overview of the 100 solution-trees, where each row represents a coding-LLM and each column corresponds to a run of the solution-seeking process. This visualization provides a high-level view of the performance of the five LLMs (\textbf{R3.2}). Specifically, 

\begin{enumerate}[topsep=2pt, partopsep=0.01pt, leftmargin=0.5cm, labelsep=0.1cm, itemsep=-2pt]
    \item \llmthree{} frequently generates buggy nodes, as indicated by the large red portions in the corresponding pie charts. Even its functional nodes perform poorly, reflected by the high average RMSE shown in the glyph in Fig.~\ref{fig:root} (right).

    \item The code generated by \llmtwo{} often takes longer to execute, as shown by the glyph on the right of Fig.~\ref{fig:root}. In each row, the pie charts are sorted by execution time, with the outer blue arc increasing from left to right. Notably, the last two trees of \llmtwo{} run especially long.

    \item \llmfour{} and \llmfive{} produce fewer buggy nodes (larger green portions in the pies) and their code generally runs faster. However, their average RMSE values are higher than those of \llmone{} and \llmtwo{}, though with noticeably lower variance, as shown in Fig.~\ref{fig:root} (right).
\end{enumerate}

\setlength{\belowcaptionskip}{-10pt}
\begin{figure*}[tb]
  \centering \includegraphics[width=\textwidth, alt={A line graph showing paper counts between 0 and 160 from 1990 to 2016 for 9 venues.}]{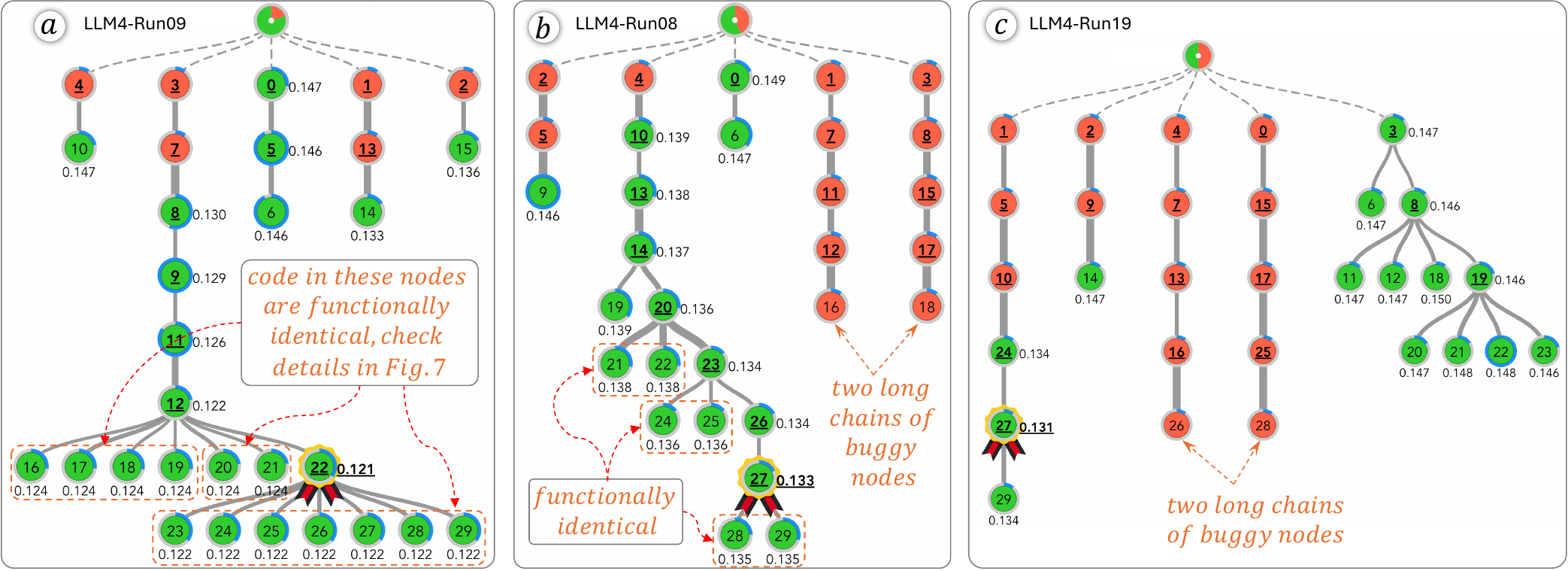}
  \vspace{-0.2in}
  \caption{The last three trees from the \llmfour{} row of Fig.~\ref{fig:ted}. The last two trees (b, c) are in the same cluster and similar but different from the tree in (a).}
  \label{fig:twotree}
\end{figure*}
\setlength{\belowcaptionskip}{0pt}

When sorting the solution-trees by structural similarity, clear clusters emerge within each LLM (Fig.~\ref{fig:ted}). This clustering provides valuable guidance for exploring and comparing the trees (\textbf{R2.2}). For example, the 20 trees from \llmthree{} form two distinct clusters, represented by purple and yellow in the dendrogram.
To understand what makes the trees in the purple cluster similar, we randomly select one for closer inspection. As shown in Fig.~\ref{fig:buggy}, nodes $0{\sim}4$ (the first $m{=}5$ nodes) are all buggy \texttt{drafts}. When generating node 5, the agent attempts an \texttt{improve} operation. However, due to the absence of any functional nodes, it defaults to generating another \texttt{draft}, following the coding policy in Fig.~\ref{fig:strategy}. This pattern continues, i.e., nodes 6, 7, 10, and many subsequent \texttt{improve} attempts also result in \texttt{draft} nodes for the same reason. As a result, the tree becomes very wide, since all \texttt{draft} nodes appear in the second level.
Many other trees generated by \llmthree{} exhibit similar structures, falling into the purple cluster. In contrast, the yellow cluster contains trees with more functional nodes. These trees tend to be deeper and structurally distinct from the shallower ones in the purple cluster.

In the \llmfour{} row of Fig.~\ref{fig:ted}, the two trees in the yellow cluster stand out from the rest. As indicated by their pie charts, both contain more buggy nodes. Fig.~\ref{fig:twotree} displays the last three trees in the \llmfour{} row. The final two (Figs.~\ref{fig:twotree}b and \ref{fig:twotree}c), belonging to the yellow cluster, contain two long chains of buggy nodes, making them structurally similar to each other but clearly different from the tree in Fig.~\ref{fig:twotree}a.

\subsubsection{Process-Level and Code-Level Comparison}
To examine the agent's coding behavior, we randomly selected one solution-tree from \llmfive{} for a detailed analysis (\textbf{R2.1}). Among the five \texttt{drafts} shown in Fig.~\ref{fig:system1}a, three (nodes 0, 2, and 4) contain buggy code. However, the agent was able to resolve these issues in a single \texttt{debug} iteration. By selecting a buggy parent node and its functional child, we can compare their code to understand how the agent fixed the bug (\textbf{R1.2}). Specifically, the issues in nodes 0 and 2 stem from the use of the \texttt{xgboost\allowbreak.XGBRegressor} package, which is unavailable in the Python environment. The agent addressed this by replacing it with \texttt{sklearn\allowbreak.ensemble\allowbreak.GradientBoostingRegressor}. The bug in node 4 was caused by the use of the \texttt{lightgbm} package, which was also not supported; the agent resolved it by switching to \texttt{sklearn\allowbreak.ensemble\allowbreak.RandomForestRegressor} in node 9.

When the agent decided to \texttt{improve} a node, it selected node 1—the best-performing node at the time—and generated node 5. Continuing downward, we observe additional \texttt{improve} attempts on the functional node 5, but these resulted in buggy nodes 6, 7, and 8. By examining and comparing the code of these nodes with that of their parent, we found that they attempted to replace \texttt{sklearn\allowbreak.ensemble\allowbreak.GradientBoostingRegressor} (in node 5) with either \texttt{xgboost\allowbreak.XGBRegressor} or \texttt{catboost\allowbreak.CatBoostRegressor}. The agent had already tried \texttt{xgboost\allowbreak.XGBRegressor} in nodes 0 and 2, which led to failures. However, because these failed attempts were neither remembered by the agent nor included in the prompt for subsequent \texttt{improve} actions, it repeated the same bug. This highlights a key limitation of the agent, \textit{it can repeat the same bug due to the lack of persistent memory}.

Through successive \texttt{debug} and \texttt{improve} iterations, the agent ultimately generates $N{=}30$ solution nodes (Fig.~\ref{fig:system1}a). The rich visual encoding supports user exploration of the tree (\textbf{R1.1}, \textbf{R2.1}). For instance, the link between nodes 8 and 11 is much thinner than the link between nodes 0 and 13, indicating fewer code changes. The award ribbon allows users to quickly identify the best-performing node—node 28. This node's code takes relatively longer to execute, as indicated by the fuller outer blue arc, though it still runs faster than node 17, which has the longest execution time. The coding plan and analysis report in Figs.~\ref{fig:system1}b and \ref{fig:system1}d help articulate the agent’s reasoning and evaluation processes.
Similar analyses can be conducted on other trees as well.

\subsubsection{LLM-Level and Code-Level Comparison}
The \projectionview{} offers an overview of the differences in code generated by the five LLMs (\textbf{R3.1}). The distinct cluster separation in Fig.~\ref{fig:system2}a suggests that the five coding-LLMs exhibit noticeably different behaviors. To further analyze these differences, we use a separate LLM to directly compare pairs of code clusters. Below are brief summaries of these comparisons (full outputs are provided in the Appendix):
\begin{itemize}[topsep=2pt, partopsep=0.01pt, leftmargin=7pt, labelsep=0.1cm, itemsep=-2pt]
    \item \llmfour{} v.s. \llmone{} (Fig.~\ref{fig:system2}-a2): \textit{..., the first collection has a direct, less automated approach focusing mostly on numeric data and standard models. In contrast, the second collection leverages more sophisticated data transformation, feature engineering, and a variety of combined models to potentially achieve better predictive performance, optimized via thorough cross-validated grid searches.
    }
    \item \llmfour{} v.s. \llmtwo{} (Fig.~\ref{fig:system2}-a3): \textit{..., the first collection leverages more powerful, tree-based ensemble methods with manual tuning and model assessment strategies, whereas the second streamlines the modeling process using linear regression techniques with a strong emphasis on pipeline integration and automatic hyperparameter optimization.
}
    \item \llmfour{} v.s. \llmfive{} (Fig.~\ref{fig:system2}-a4): \textit{... Both collections handle preprocessing of data, model training, validation, and prediction output. However, the second collection adopts more complex, layered machine learning strategies involving blending and stacking of models to potentially enhance prediction accuracy significantly over diverse single-model approaches used in the first collection.
}
\end{itemize}

These summaries help contrast the behaviors of different coding-LLMs and shed light on why one may outperform another. For example, \llmone{} produces code with a lower RMSE than \llmfour{}—as indicated by the glyphs in Fig.~\ref{fig:root}—because it ensembles multiple models and performs comprehensive parameter tuning via grid search.

The \packageview{} provides concrete evidence to support the summaries (\textbf{R3.1}). For example, the dominant use of \texttt{lightgbm} in \llmthree{} (Fig.~\ref{fig:system2}-b1) supports the summary that \llmthree{} is biased toward \texttt{lightgbm} and lacks flexibility in adopting other models.
When comparing \llmfour{} with \llmone{}, the summary notes that \llmone{} tends to ensemble models and use cross-validation for parameter tuning. By sorting packages based on their usage in \llmone{}, we observe that \texttt{sklearn\allowbreak.model\_selection\allowbreak.cross\_val\_score} appears significantly more frequently (Fig.~\ref{fig:system2}-b2). This package is commonly used for cross-validation, reinforcing the earlier comparative summary.
Sorting by \llmfive{}, we found that it uses the package \texttt{os} much more often (Fig.~\ref{fig:system2}-b3).
Inspecting the details in the \codeview{}, \llmfive{} frequently uses \texttt{os.makedirs("./working",\allowbreak exist\_ok=True)} to proactively create a directory before saving outputs. In contrast, other LLMs assume the directory already exists and write outputs into it without verification. This suggests that \llmfive{} adopts a more cautious and robust coding style than the others.

The \packageview{} also reveals each LLM’s preference for different packages when implementing the same functionality. For example, \llmtwo{} often uses \texttt{sqrt()} from the \texttt{numpy} package to compute square roots, whereas other LLMs prefer \texttt{sqrt()} from the \texttt{math} package. The package sorting feature makes such differences easy to identify.

\subsection{Actionable Insights for AIDE Improvements}
\label{sec:insight}
The contribution of our work does not lie in creating novel visualizations; rather, it lies in coordinating existing ones to better organize information and help domain experts uncover insights that are difficult to obtain through their routine workflows. To emphasize this point, we showcase unique insights derived from our system that have directly guided improvements to AIDE.
As explained in Sec.~\ref{sec:background}, the two core components of AIDE are the LLM backbone and the coding policy. Our system provides actionable insights into both.

\subsubsection{Repeating the Same Bug Between Iterations}
While exploring different trees and examining the execution output of buggy nodes, the experts observed that the agent may repeat the same bug across iterations. For example, in Fig.~\ref{fig:twotree}b, node 9 fixes the bug in node 5 caused by the unavailability of \texttt{xgboost}. However, in node 17 (generated to debug node 15), the agent again attempts to import \texttt{xgboost}.
Ideally, since nodes 5 and 9 are generated before node 17, the agent should be aware of the unavailability of \texttt{xgboost} and know its appropriate substitute. However, because the \texttt{debug} process focuses solely on the immediate parent buggy node (see the prompt for \texttt{debug} in~\cite{aideprompt}) and lacks memory of previous debugging history, the agent fails to recognize that it is repeating earlier mistakes.
More concerningly, we observed cycles of repeated bugs: in one case, the second iteration fixed bug $A$ from the first iteration but introduced bug $B$; then, in attempting to fix $B$, the third iteration reintroduced bug $A$, resulting in a cycle. These repeated bugs do not contribute new knowledge to the agent but instead consume valuable exploration budget.

The experts' routine workflow, even with the aid of the visualization tool in Fig.~\ref{fig:aidevis}, is hard for them to easily identify buggy nodes, determine their root causes, and trace recurring bugs across a solution-tree. As a result, the experts were unaware of the recurring bug issue. Revealing this problem helped the experts better design their LLM prompt to improve the agent. In this particular case, they have proposed maintaining a bug list and including it in the prompt for \texttt{debugging}, instructing the LLMs to avoid known bugs. A separate research project has been initialized to comprehensively evaluate this improvement.

\subsubsection{Wasting Computational Resources}
Another surprising finding comes from the function-level similarity analysis of the generated code. As shown in Figs.~\ref{fig:matrix} and~\ref{fig:twotree}a, child nodes created to \texttt{improve} the same parent node can produce functionally-identical code. This occurs because these \texttt{improve} attempts originate from a shared base node and are guided by similar prompts. Although later attempts can access previously generated code, there is no explicit enforcement of diversity. This issue is particularly pronounced in \llmfour{} (e.g., Figs.~\ref{fig:twotree}a and  \ref{fig:twotree}b). Repeatedly executing functionally-identical code does not enhance the agent’s performance and merely wastes computational resources. In a commonly used AIDE benchmark—Chan et al.\cite{chan2024mle}—the timeout for each node is set to 9 hours, meaning each redundant execution could waste up to 9 hours.

Without the proposed function-level code comparison, this issue is difficult to detect and easily overlooked, as it does not break the solution-seeking process. However, once revealed, the wasted computational resources became a significant concern for the experts. As a remedy, they proposed that when \texttt{improving} a node, the LLM should be instructed to diversify from the code of the current node’s \textbf{\textit{siblings}}. Similarly, during \texttt{drafting}, previously generated \texttt{draft} solutions should be made visible to the agent to promote diversity from the outset. These changes aim to guide the LLM to explore underrepresented regions of the solution space.

\subsubsection{Greedy Coding Policy}
The enriched visual encoding of the \treeview{} helps experts analyze the agent’s coding policy and identify its limitations. For instance, the experts initially overlooked the greedy nature of the \texttt{improve} operation, i.e., always selecting the best-performing node to \texttt{improve}. However, after observing many wide subtrees—caused by repeated improvements on the same node without any performance gain—they realized that the agent could get trapped in a local minimum within the solution space.
For example, in Fig.~\ref{fig:twotree}a, nodes 23–29 are all \texttt{improving} the same base node, 22, as it was the best-performing node at the time. Yet, none of these improvements led to measurable gains. 
Furthermore, the metric values for nodes 16–29 are very close ($0.121{\sim}0.124$), and all are computed on the validation set. As a result, small differences between them may not be meaningful, and a node that appears second-best on validation data could potentially perform better on the test set.

These observations immediately alerted the experts and prompted them to refine the greedy improvement policy by considering alternative nodes instead of focusing solely on the best one. Specifically, they now select a node to \texttt{improve} by sampling from a probability distribution constructed using a \textit{softmax} over all nodes’ metric values. This approach ensures that the best-performing node still has the highest probability of being chosen, while also giving other nodes a chance—enabling the agent to escape local minima in the solution space. In certain applications, this adjustment—especially with careful tuning of the \textit{softmax} temperature—has led the agent to generate better solutions. The experts are conducting further experiments to thoroughly evaluate this adjustment in the coding policy.

\subsection{Domain Experts' Feedback}
\label{sec:feedback}
Following the \textit{guided exploration+think-aloud discussion} protocol, we conducted case studies with the five experts ($E_1{\sim}E_5$). Each expert explored multiple cases, interacted with the system, and provided feedback at the end. This section summarizes their feedback, which primarily focused on three aspects: (1) facilitating the understanding of coding agents, (2) generating actionable insights for improving AIDE, and (3) discussing desired extensions of the system.

Through extensive exploration of different trees, $E_3$ and $E_5$ developed a deep understanding of various tree formations. Both experts found our visual encoding more informative than the original AIDE visualization, as it allowed them to better interpret how coding policies influence tree structures. For example, they noted that since the \texttt{improve} operation always extends the best-performing node, a node that outperforms its parent results in the creation of a new tree level. Based on this, $E_3$ concluded that tree depth can serve as an indicator of progress in the solution-seeking process, with deeper trees reflecting greater progress.
Additionally, $E_3$ appreciated the system’s ability to provide an overview of a large number of solution-trees (Fig.~\ref{fig:root}). He also stated that \llmthree{} may not necessarily be inferior to other LLMs, but rather less flexible, and pointed out that if \texttt{lightgbm} were available, much of \llmthree{}'s code could become functional, potentially leading to better performance.
All five experts found the comparison feature in the \codeview{} particularly valuable, as it effectively highlighted code changes over iterations. In conjunction with the LLM-level comparison, $E_1$ and $E_3$ summarized two major improvement strategies employed by the agent: (1) exploring different ML models and ensembling them, and (2) using fewer ML models while focusing on feature engineering and hyperparameter tuning. $E_4$ found this especially intriguing, noting that these two strategies align closely with how humans improve models.

The experts especially appreciated that many insights from our system could lead to actionable improvements, such as those exemplified in Sec.~\ref{sec:insight}. These insights are typically invisible in their routine workflow, in which the experts focused more on comparing the generated solutions based solely on their numerical performance values. The details uncovered by the visualizations, e.g., a chain of nodes with repeating bugs and functionally-identical code in a wide sub-tree, were highly insightful and prompted immediate actions towards improving the framework. $E_1$ and $E_2$ extensively use AIDE in ML model building and heavily rely on the tree structure visualization shown in Fig.~\ref{fig:aidevis}. However, they have limited flexibility to check the execution result of each node; at times, they even need to re-run the code to reproduce the execution output for further investigation. The information is fragmented and scattered, hindering them from effectively analyzing the solution-seeking process—let alone identifying potential improvements. The coordinated visualization system consolidates all relevant information, and its intuitive interactions help them focus on analyzing generated solutions and enhancing the AIDE framework.

Additionally, the experts recommended several desired features to enhance our system.
\textit{First}, $E_1$ and $E_2$ emphasized the need to visualize the prompts for the three operations in the coding policy. This would help them better understand what the agent knows when generating code. For example, the agent may unknowingly repeat previous bugs because earlier buggy nodes were not included in the prompt for code generation. Externalizing the prompts would allow experts to better assess the agent’s limitations and refine prompt design accordingly.
\textit{Second}, $E_5$ proposed a new scatterplot with more meaningful axes for visualizing the generated code. While similar to the scatterplot in Fig.~\ref{fig:system2}a, this new visualization would position points (representing code) based on key metrics such as execution time, performance score, and code length. Such a representation would facilitate code comparison and reveal broader patterns across different LLMs.
\textit{Third}, $E_1$, $E_4$, and $E_5$ found the \packageview{} very intuitive and effective. They also discussed the potential for incorporating package co-occurrence analysis, i.e., the joint appearance of two packages.
The above feedback provides valuable directions for further improving our system.

\section{Discussion, Limitations, and Future Work}
\label{sec:discussion}
We would like to highlight several known limitations, discuss their implications, and suggest possible improvements for the near future.

First, our work focuses on AIDE as a representative example to support general tree-based agentic coding frameworks. Although we concentrate solely on AIDE, we see no major obstacles in adapting our system to other tree-based coding frameworks. Specifically, since any such framework produces a solution-tree, our \treeview{} can be directly applied to visualize it, with only minor adjustments to the visual encoding if necessary.
The ultimate output of any coding agent is a collection of code. Our \codeview{} and \projectionview{} are designed to be general-purpose and can visualize, compare, and cluster any collection of code for insight discovery.
The \packageview{}, however, may face challenges when analyzing code written in languages other than Python. This is due to our assumption that a piece of code’s behavior can be inferred from its imported packages—an assumption that may not hold in other programming languages. While we acknowledge this limitation, we remain optimistic about the generalizability of our system, as it targets agents for ML modeling tasks—a domain in which Python is the most widely used language.

Second, part of the overview presented in Fig.~\ref{fig:root} relies on evaluation metric values computed by the generated code. In Sec.~\ref{sec:casestudy}, the evaluation metric used is RMSE, and the values produced by the functional code are correctly computed. However, $E_1$ observed that in some scenarios, the generated code does not produce correct metric values. In one task, for example, he intended to use top-$5$ recall on the validation data as the evaluation metric. However, some generated code versions computed it using the training data, while others failed to compute it correctly—yet still returned a numerical value as the metric.
Incorrect metric values can mislead the coding policy, causing it to select suboptimal nodes to \texttt{improve}. A simple remedy is to allow users to provide standalone code for calculating the metric, making the evaluation more objective.

Third, in function-level code comparison, there are likely many more corner cases that our current solution does not yet address. The six cases in Fig.~\ref{fig:codediff}a–f are those we observed during our exploration, and we believe this list will continue to grow as we apply our system to compare more code examples. 
For instance, if the \texttt{set} structure is used in both versions of the code and contains the same elements, but in different orders, the functionality remains unchanged.
This case, among many others, will be added to our list to make the function-level code comparison more robust in the future.

Lastly, we plan to use our system to conduct case studies with more LLMs to profile their coding behaviors. The resulting comparative insights could be highly valuable for users who are uncertain about which LLM to choose from the growing number of available options.
\section{Conclusion}
In this paper, we introduced a visual analytics system for exploring the solution-seeking process of LLM-powered coding agents. We began by enhancing an existing visualization tool with enriched visual encodings. Next, we proposed a three-level analysis framework to better understand these agents, focusing on comparisons of their generated code, solution-seeking processes, and underlying LLMs. Finally, we conducted case studies in which the agents solved real-world Kaggle competition tasks, allowing us to compare their behaviors in practice. The insights gained from these studies, along with feedback from domain experts, validate the effectiveness of our system.




\bibliographystyle{abbrv-doi-hyperref}

\bibliography{template}








\end{document}